\documentclass{article}

\usepackage[english]{babel}

\usepackage[letterpaper,top=2cm,bottom=2cm,left=3cm,right=3cm,marginparwidth=1.75cm]{geometry}

\usepackage{amsmath}
\usepackage{amsfonts}
\usepackage{graphicx}
\usepackage{parskip}
\usepackage{comment}
\usepackage{csquotes}
\usepackage[colorlinks=true, allcolors=blue]{hyperref}
\usepackage[
backend=biber,
style=authoryear,
]{biblatex}
\usepackage{subfig}
\usepackage{booktabs}
\usepackage[colorlinks=true, allcolors=blue]{hyperref}
\usepackage{mathtools}

\newtheorem{theorem}{Theorem}[section]

\def\sym#1{\ifmmode^{#1}\else\(^{#1}\)\fi}

\title{Power-Law Trends in Speedrunning and Machine Learning}
\author{Ege Erdil \and Jaime Sevilla\footnote{Ege Erdil ideated the project, conducted all the analysis and wrote the first draft, Jaime Sevilla supervised and edited the paper.}}
\date{%
    Epoch \hspace{0.15cm}%
}
\begin{document}
\maketitle

\begin{abstract}
We find that improvements in speedrunning world records  follow a power-law pattern. Using this observation, we answer an outstanding question from \cite{Sevilla2021}: How do we improve on the baseline of predicting no improvement when forecasting speedrunning world records out to some time horizon, such as one month? Using a random effects model, we improve on this baseline for relative mean \( L^2 \) error made on predicting out-of-sample world record improvements as the comparison metric at a \( p < 10^{-5} \) significance level. The same set-up improves \textit{even} on the ex-post best exponential moving average forecasts at a \( p = 0.15 \) significance level while having access to substantially fewer data points. We demonstrate the effectiveness of this approach by applying it to Machine Learning benchmarks and achieving forecasts that exceed the baseline. Finally, we interpret the resulting model to suggest that 1) ML benchmarks are far from saturation and 2) sudden large improvements in Machine Learning are unlikely but cannot be ruled out.
\end{abstract}

\section{Introduction}

Forecasting record progressions is a problem of significant interest in many domains. This problem is of particular interest for the future of AI for two reasons:

\begin{enumerate}
\item Records attained on machine learning benchmarks are in this class, and since many different metrics are employed in machine learning, we want a fairly general method to forecast them.
\item It's possible that we'll see discontinuities in AI development. Indeed, this seems to be central to the difference between the ``slow takeoff" and ``hard takeoff" pictures as explained in \cite{Christiano2018}. Observing how discontinuous progress tends to happen could help us better understand which AI development scenarios are more likely.\footnote{Ideally, we'd get data about AI development directly, but this data is difficult to access in practice.}
\end{enumerate}

Directly studying machine learning benchmarks is valuable but difficult. Machine learning benchmarks are measured in many different ways. For example, scores in games, the fraction of tasks on a list that an AI can solve, the number of objects correctly classified in a classification problem, \textit{et cetera}. The diversity of metrics makes it difficult to unify the results. In addition, even the most popular benchmarks don't report many world record improvements: the most improved upon benchmarks cap out at around 15 improvements to the previous world record.

\begin{figure}[h]
\centering
\includegraphics[width=1\textwidth]{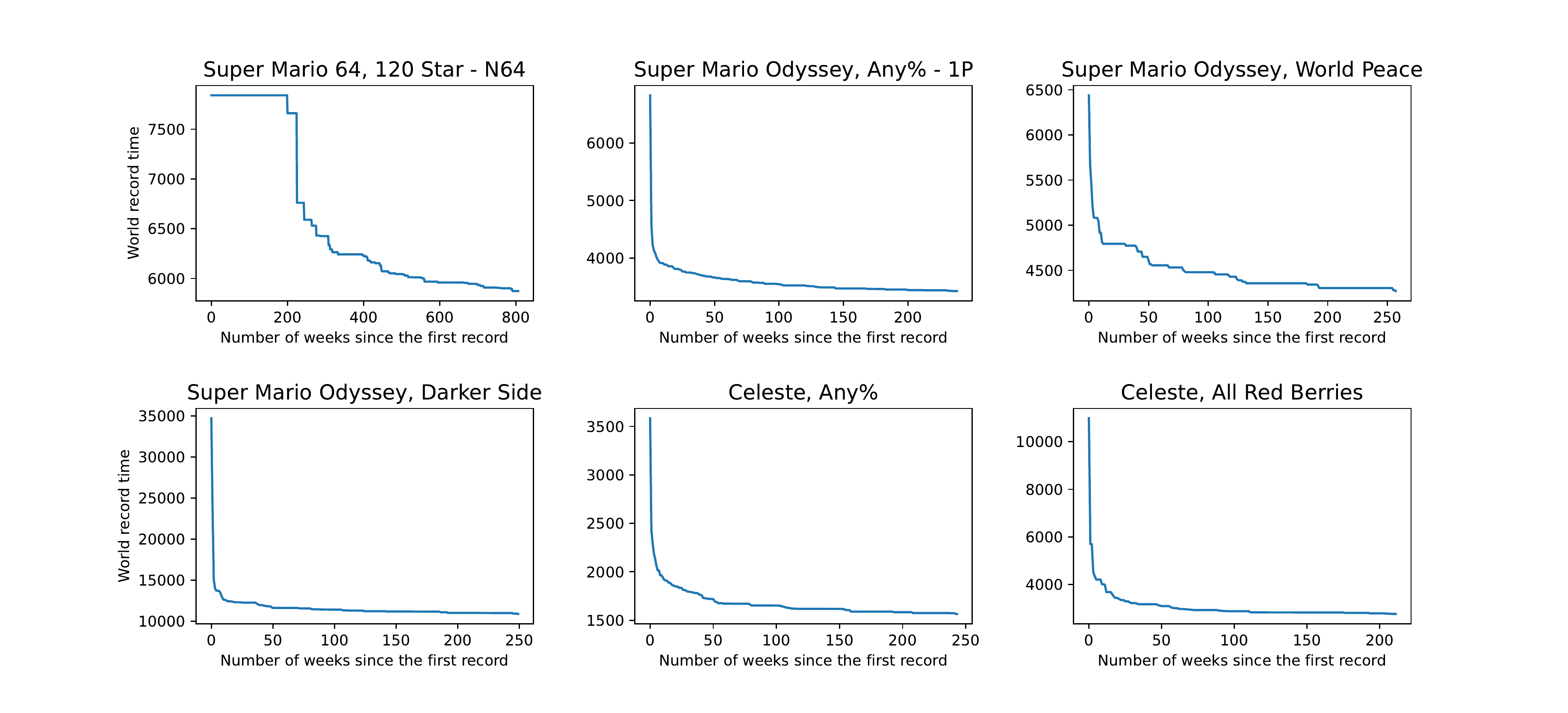}
\caption{\centering \small World records times (in seconds) on six different speedrunning categories over time. Data was obtained from \href{https://www.speedrun.com/}{speedrun.com}, a site that provides leaderboards, resources, forums and more, for speedrunning. Our dataset spans a total of 25 speedrunning categories and 1731 world records.}
\label{fig:sample_plot}
\end{figure}

\cite{Sevilla2021} observes that none of these problems come up with speedruns. Speedrunning is a competitive activity in which the goal is to finish a computer game in the shortest amount of time possible while abiding by some constraints agreed upon by the community.\footnote{ Each set of constraints can make for a different ``category" of speedrunning for the same game, and we treat them separately in our analysis.} The website \href{https://www.speedrun.com/}{www.speedrun.com} is dedicated to storing speedrunning world records with a publicly accessible API. While not exhaustive, leaderboards on this page have become well-known and reputable, and as a result WR data after 2010 or so is quite reliable. Using this, we can compile a large dataset of world records and analyze how these change over time. A sample speedrunning world record time series is given in Figure \ref{fig:sample_plot} for the reader's convenience.\footnote{Code to reproduce the results in this write-up is available at \href{https://colab.research.google.com/drive/1huIHSpkbJggHd0khO_di6-yGl5hyFkWf?usp=sharing}{this Colab notebook}.}

We can split the problem of forecasting such a time series into two components:

\begin{itemize}
\item Predicting \textit{when} the next world record improvement will occur.
\item Predicting the \textit{size} of the next world record improvement.
\end{itemize}

If we could accurately predict both the timing and sizes of world records, we could predict the world record time expected by each date in the future.

We'll have less to say about the timing of records going forward; most of our analysis will be focused on predicting improvement size. Still, relatively simple regression models for timing can be combined with our findings about predicting sizes to settle an open problem from \cite{Sevilla2021} of beating the naive baseline in forecasting what the world record will be at time \( t + \Delta t \) using only data up to time \( t \) for some values of \( \Delta t \), e.g., four or eight weeks. We explain in detail the strategy we used in section \ref{ssec:time-series-forecasting}.

We will then show in section \ref{sec:ml-benchmarks} how to apply similar methods to predicting state-of-the-art (SOTA) improvements in ML benchmarks. In particular, we will show how a random effects model with power-law decay can be used to beat a no-improvements-predicted baseline.

Finally, in section \ref{sec:interpretation}, we will discuss possible interpretations of the models. In particular, we will show how the ML model suggests that ML benchmarks are not yet saturated and that sudden large improvements are relatively unlikely, though they cannot be ruled out.

We refer readers interested in previous work in record forecasting to the literature review included in \cite{Sevilla2022}. To our knowledge, this is the second piece of academic work after \cite{Sevilla2021} to study videogame speedrunning records, and the first article to study forecasting of Machine Learning benchmarks.

\section{Power-law trends in speedrunning}

\begin{figure}[h!]
    \centering
    \subfloat[\centering Relative improvements in the world record time of \textit{The Legend of Zelda: Breath of the Wild -- any \%} category against the number of world records that have occurred so far. ]{{\includegraphics[width=0.45\textwidth]{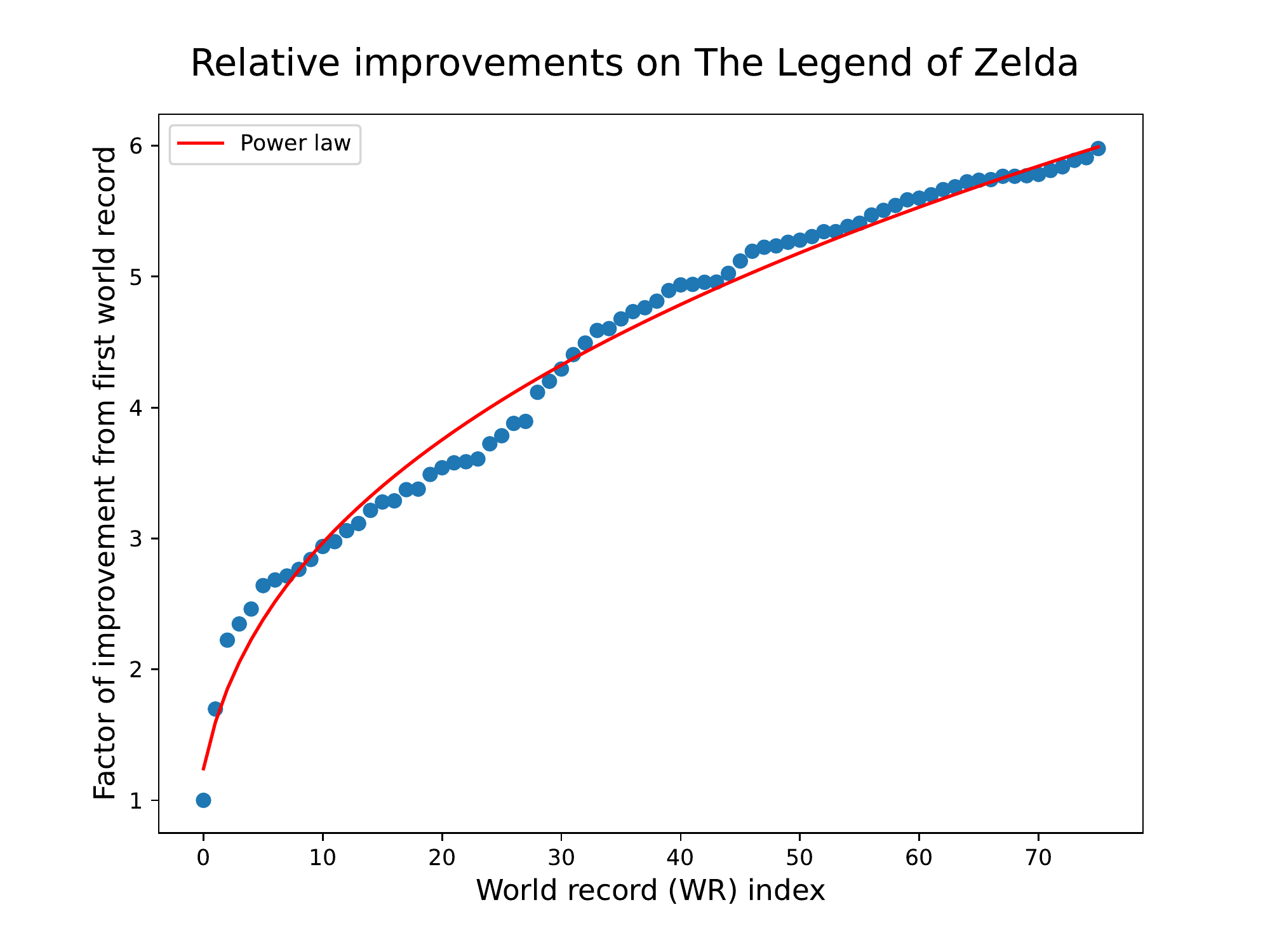} }}%
    \qquad
    \subfloat[\centering Mean double logarithm of world record improvements across all categories against the number of world records before the improvement. Shaded areas show standard deviations across 25 categories. ]{{\includegraphics[width=0.45\textwidth]{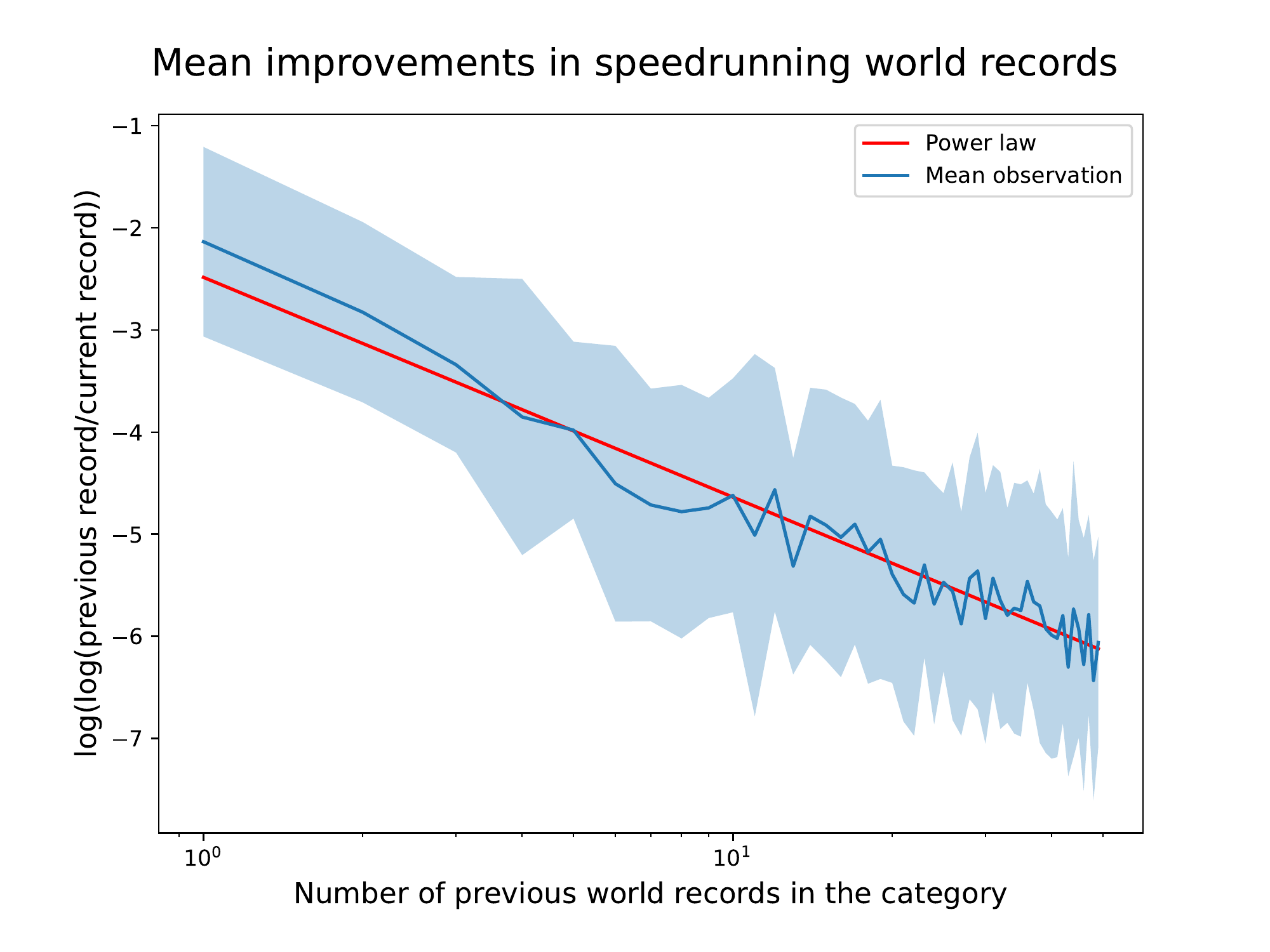} }}%
    \caption{\centering \small A clear and homoskedastic power-law trend is apparent in the data, for individual series, such as \textit{Zelda} (Left), and for the full dataset (Right).}%
    \label{fig:power_law_plot}%
    \label{fig:zelda_plot}
\end{figure}

When it comes to the size of the improvements, a potentially useful trend had been noticed earlier in \cite{SevillaAddendum}: World record improvements in speedrunning seem to be closely related to power-laws, and power-laws can be used to improve on the naive baseline forecast of ``the world record will not change." An example of how the power-law fit works in practice is given in Figure \ref{fig:zelda_plot}.

Before imposing any particular model on the data, let's average relative improvements in the world record at a particular world record (WR) index over all 25 categories in our dataset. This gives us the remarkable plot in Figure \ref{fig:zelda_plot}.


The power-law fits remarkably well: variance across categories remains roughly stable as we move towards higher world record indices. This is the main regularity we exploit in improving over \cite{Sevilla2021} and constructing a new random effects model to forecast world record improvements in speedrunning.

\section{Predicting record improvements in speedrunning}

In this section, we'll outline a model that uses the power-law decay insight from the introduction to forecast future world record improvements in speedrunning. We'll then benchmark this model on a dataset by having it perform out-of-sample predictions and comparing these predictions with other baseline methods, including always predicting no improvement and predicting an exponential moving average of past world record improvements.

The main result is that the power-law insight, when combined with an appropriate model fitting method that properly uses data parallelism across speedrunning categories, allows us to beat naive baseline methods in forecasting future world record improvements.

\subsection{Data}

The dataset consists of all categories of the 100 most popular games on \href{https://www.speedrun.com/}{speedrun.com} with 50 or more world records as of October 14, 2022. This amounts to a total of 25 categories and 1731 world records \footnote{We repeated the analysis with a more expansive dataset of 53 categories with 50 or more world records among the 1000 most popular games. This didn't significantly change the results in this section.}$^{,\thinspace}$\footnote{The data used is fetched using the same script from \cite{Sevilla2021}, though we use an updated version of his dataset retrieved on October 14, 2022. You can download the dataset at \href{https://epochai.org/data/speedrunning}{epochai.org/data/speedrunning}.
}


\subsection{Model}

There are two main features apparent in the data that the ideal model needs to capture:

\begin{enumerate}
    \item The diminishing returns structure of speedrun records exhibited in Figure \ref{fig:zelda_plot}. Big improvements become rarer over time, and the typical improvement gets smaller over time, in a way well-described by power-laws.
    \item Discontinuities in rates of progress, or ``trend breaks." Qualitatively, the data for some speedruns shows clear breaks from previous trends where a speedrun that used to be seeing slow progress suddenly switches to a mode of faster progress. See Figure \ref{fig:zelda_plot} for an example.
\end{enumerate}

The second feature is a problem beyond the scope of this work, though we think it'd be interesting for further work to study this phenomena.\footnote{ Testing any model of trend breaks is complicated by the small amount of them found in the dataset: despite us having 25 time series each with more than 50 world records, only a few of these time series show any trend breaks at all, and there are no time series that qualitatively appear to show two or more trend breaks. So when it comes to trend breaks, we're operating in a data-scarce environment, making it difficult to test any models that we might come up with.}

To properly model the first feature, we experimented with many models, some of which were quite complicated.\footnote{The primary model forms we experimented with were Hidden Markov Models (HMMs). However, when fit properly to the data, they tended to collapse down to determinism, and their performance depended strongly on the exact law of motion assumed for the latent variable. This meant that a simple regression model does as well or better without the accompanying complexity of dealing with HMMs.} However, the best-performing model we could produce and fit was a relatively simple random effects model.

The model looks like this: We assume that for each category \( c \), there are category-specific coefficients \( \alpha_c, \beta_c \) such that we have

\[ \log \log \left( \frac{R_{c, t}}{R_{c, t+1}} \right) = \alpha_c + \beta_c \times \log(t) + \varepsilon_{c, t} \]

where \( R_{c, t} \) is the \( t \)th record in category \( c \) and the \( \varepsilon_{c, t} \sim N(0, \sigma^2) \) are independent and identically distributed normal error terms. Because the model is a random effects model, the coefficients \( \alpha_c, \beta_c \) are independent and normally distributed random variables.

This model is appealing in its simplicity and has the advantage that fitting it is relatively straightforward by making use of Python's \textit{statsmodels} package \parencite{seabold2010statsmodels}.

\subsection{Results}

We test the model by fitting it to the first \( K = 10 \) world records in each category and using the fitted model to predict subsequent relative improvements in the world record up to the end of each time series. The results are presented in Table \ref{tab:reg-results}.

\begin{table}[h]
\caption{Mixed Linear Model Regression Results: Speedrunning}
\begin{center}
\begin{tabular}{llll}
\hline
Model:            & MixedLM & Dependent Variable: & \( \log \log \left( \frac{R_{c, t}}{R_{c, t+1}} \right) \)  \\
No. Observations: & 225     & Method:             & ML            \\
No. Groups:       & 25      & Scale:              & 0.7964        \\
Min. group size:  & 9       & Log-Likelihood:     & -314.9688     \\
Max. group size:  & 9       & Converged:          & Yes           \\
Mean group size:  & 9.0     &                     &               \\
\hline
\end{tabular}
\end{center}

\begin{center}
\begin{tabular}{lrrrrrr}
\hline
                          &  Coef. & Std.Err. &       z & P$> |$z$|$ & [0.025 & 0.975]  \\
\hline
\( \mathbb E[\alpha_c] \)                & -2.029 &    0.149 & -13.641 &       0.000 & -2.321 & -1.738  \\
\( \mathbb E[\beta_c] \)              & -1.297 &    0.110 & -11.750 &       0.000 & -1.514 & -1.081  \\
\( \operatorname{var}(\alpha_c) \)              &  0.075 &    0.090 &         &             &        &         \\
\( \operatorname{cov}(\alpha_c, \beta_c) \) &  0.000 &    0.000 &         &             &        &         \\
\( \operatorname{var}(\beta_c) \)           &  0.112 &    0.054 &         &             &        &         \\
\hline
\end{tabular}
\end{center}
\label{tab:reg-results}
\end{table}

We then compare the random effects model with a few other choices:

\begin{itemize}
    \item A \textbf{naive baseline} that always predicts zero relative improvement in the world record.
    \item A \textbf{fixed effects model} that is fit to data only from a single category. This model has the same regression equation as the random effects model. The difference is in how parameters are shared across categories and which data the model sees during fitting.
    \item As a baseline better than predicting no improvement, we add an \textbf{exponential moving average model}. This is a natural non-parametric model to try to forecast logarithmic improvements in the world record: we look at the most recent improvements and average them using exponentially decaying weights. The decay parameter here is shared across all categories.
\end{itemize}

These models have the advantage that they get to see all data points up to and including the \( N \)th world record when forecasting the relative improvement between the \( N \)th and \( N+1 \)th world record, and they have the disadvantage that they don't share parameters across categories the way the random effects model does. The results are presented in Table \ref{tab:benchmark-results}.

\begin{table}[h]
\centering
\begin{tabular}{@{}lcc@{}}
\toprule
\textbf{Model}                                          & \textbf{Mean \( L^2 \) error} & \textbf{Mean \( L^1 \) error} \\
\midrule
Baseline: always predict zero improvement               & \( 3.332 \% \)                & \( 0.768 \% \)                \\
Random effects model                                    & \( \bf{3.203 \%} \)                & \( \bf{0.591 \%} \)                \\ 
Fixed effects model with category-specific coefficients & \( 3.221 \% \)                & \( 0.615 \% \)                \\
Ex-post best exponential moving average model           & \( 3.319 \% \)                & \( 0.678 \% \)                \\
\bottomrule
\end{tabular}
\caption{Benchmark test results.}
\label{tab:benchmark-results}
\end{table}

While the difference in mean \( L^2 \) error across models is unspectacular, the difference is in fact highly statistically significant because the difficulty of examples is correlated across models. That is, while for some observations, both models perform poorly in an absolute sense, the random effects model outperforms the naive baseline in almost every case (see Figure 3b).

\begin{figure}[h!]
    \centering
    {\includegraphics[width=0.85\textwidth]{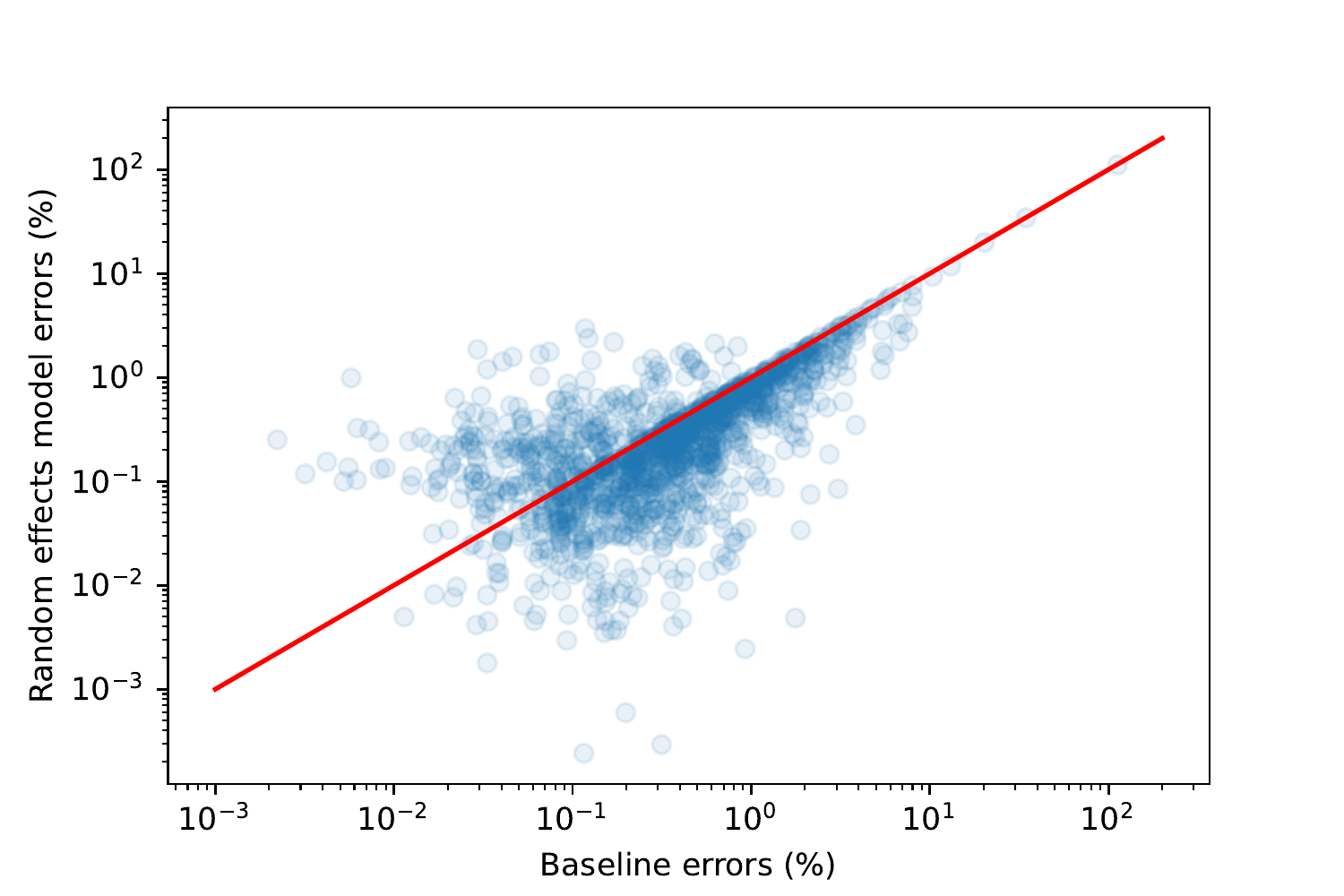} }%
    \caption{\centering \small A log-scale scatter plot of the relative errors made by the naive baseline of predicting zero improvement on all cases versus the random effects model. The large \( L^2 \) errors are driven by the few improvements in the upper right corner, which all models have trouble predicting. The red line is defined by \( x = y \). The random effects model outperforms the baseline model in X\% of the cases (the ones below the red line).}%
    \label{fig:error_scatter_plot}%
\end{figure}


This means the differences become amplified when bootstrapping: randomly sampling from the 1481 out-of-sample improvement forecasts that the models are benchmarked on shows that the random effects model and the baseline at a statistically significant \( p = 10^{-5} \) confidence threshold.

The data is too noisy to guarantee that the random effects model is superior to the ex-post best exponential moving average model, and we are only able to make this claim at a \( p = 15 \% \) confidence level, which is not statistically significant. However, the models appear at least comparable in performance. Given the more limited data access of the random effects model, even this result is quite promising.

As a final robustness check, the reader might be interested to know what happens if we pick a value of the cutoff \( K \) different from \( K = 10 \). We list the out-of-sample mean relative \( L^2 \) error of the models for a few different values of \( K \) in the table below.

\begin{table}[h]
\centering
\begin{tabular}{@{}llllll@{}}
\toprule
 & K = 3 & K=5 & K=10 & K=20 & K=30 \\ \midrule
\begin{tabular}[c]{@{}l@{}}Baseline: always predict\\ zero improvement\end{tabular} & 3.546\% & 3.314\% & 3.332\% & 1.346\% & 1.357\% \\
Random effects model & \textbf{3.285\%} & \textbf{3.163\%} & \textbf{3.203\%} & \textbf{1.224\%} & 1.272\% \\
\begin{tabular}[c]{@{}l@{}}Fixed effects model \\ with category-specific \\ coefficients\end{tabular} & 3.538\% & 3.173\% & 3.221\% & 1.225\% & \textbf{1.252\%} \\
\begin{tabular}[c]{@{}l@{}}Ex-post best exponential \\ moving average model\end{tabular} & 3.790\% & 3.447\% & 3.319\% & 1.268\% & 1.281\% \\ \bottomrule
\end{tabular}
\caption{Out-of-sample mean relative \( L^2 \) error performance of the models for different values of the cutoff \( K \). Values representing the best performance achieved are in bold for each value of \( K \).}
\label{tab:cutoff-comparison}
\end{table}

The basic finding is that the random effects model has a significant edge over other methods when \( K \) is small, but as \( K \) gets large, the edge over the fixed effects and exponential average methods disappears. This makes sense: having more data points per time series means there's less variance in the fixed effect estimators, and because of the power-law nature of the decay in world record improvement magnitudes, exponential average methods can achieve better variance reduction by averaging over more values (i.e., having a long memory) when \( K \) is large. These findings are, therefore, consistent with the power-law decay we've posited.

\subsection{Time series forecasting}
\label{ssec:time-series-forecasting}

To resolve the open problem of speedrun time series forecasting from \cite{Sevilla2021}, we use the following approach: say that \( T_{c, t} \) is the clock time at which the \( t \)th world record in category \( c \) is achieved. We use a regression model

\[ \log(T_{c, t+1} - T_{c, t}) = \zeta_c + \gamma_c \times t + \varepsilon_{c, t} \]

and fit it to the world record data \( t \leq N \) for a single category \( c \), where \( N \) is large enough for the regression to be properly identified, e.g., \( N \geq 15 \). We then check if the coefficient \( \gamma_c \) has a positive point estimate. If so, we use samples drawn from this regression model to simulate time gaps between future world records; if not, we do Monte Carlo instead by bootstrapping from the historical time gaps \( T_{c, t+1} - T_{c, t} \) up to \( t = N-1 \). This is a crude way of implementing our prior belief that the time between records should not exponentially decay over time, and it performs well empirically.

We then take the above method of simulating time gaps between successive world records and combine it with the fixed effects, category-specific power-law regression to simulate sample world record trajectories. We fix \( \Delta t \) at a particular value such as eight weeks and run one simulation starting from \( T_{c, N} \) for each category \( c \) and each record number \( 15 \leq N \leq 45 \) to predict the world record at time \( T_{c, N} + \Delta t \). The exact boundaries are not important as long as 1) we know the record at time \( T_{c, N_{max}} + \Delta t \) and 2) \( N_{min} \) is large enough to ensure that the regression estimates are not too noisy.

The results are decisively in favor of our model. The approach described above achieves a mean \( L^2 \) error of \( 3.17 \% \) compared with \( 3.99 \% \) for the baseline, and the difference is statistically significant at a \( p < 10^{-6} \) threshold at a horizon of \( \Delta t = 8 \textrm{ weeks} \). We can therefore comfortably declare that the problem set out in \cite{Sevilla2021} of doing out-of-sample time series forecasting on speedrunning data better than the naive baseline has been solved.

\section{Predicting ML benchmarks}
\label{sec:ml-benchmarks}

In this section, we apply a model similar to the one developed in the speedrunning section to a dataset of machine learning benchmarks. Similarly to the speedrunning case, we are able to improve over the naive baseline of predicting no improvement. However, the scarcity of longitudinal data makes it tricky to construct good moving average benchmarks to compare our models to.

The results about the exact type of decay in state-of-the-art (SOTA) improvements here are less impressive because ML benchmark time series generally only have a few improvements over previous state-of-the-art results. We consider both exponential decay and power-law decay, but neither data nor theory is sufficient to establish which one is more appropriate.

\subsection{Data}

\begin{figure}[h]
\centering
\includegraphics[width=1\textwidth]{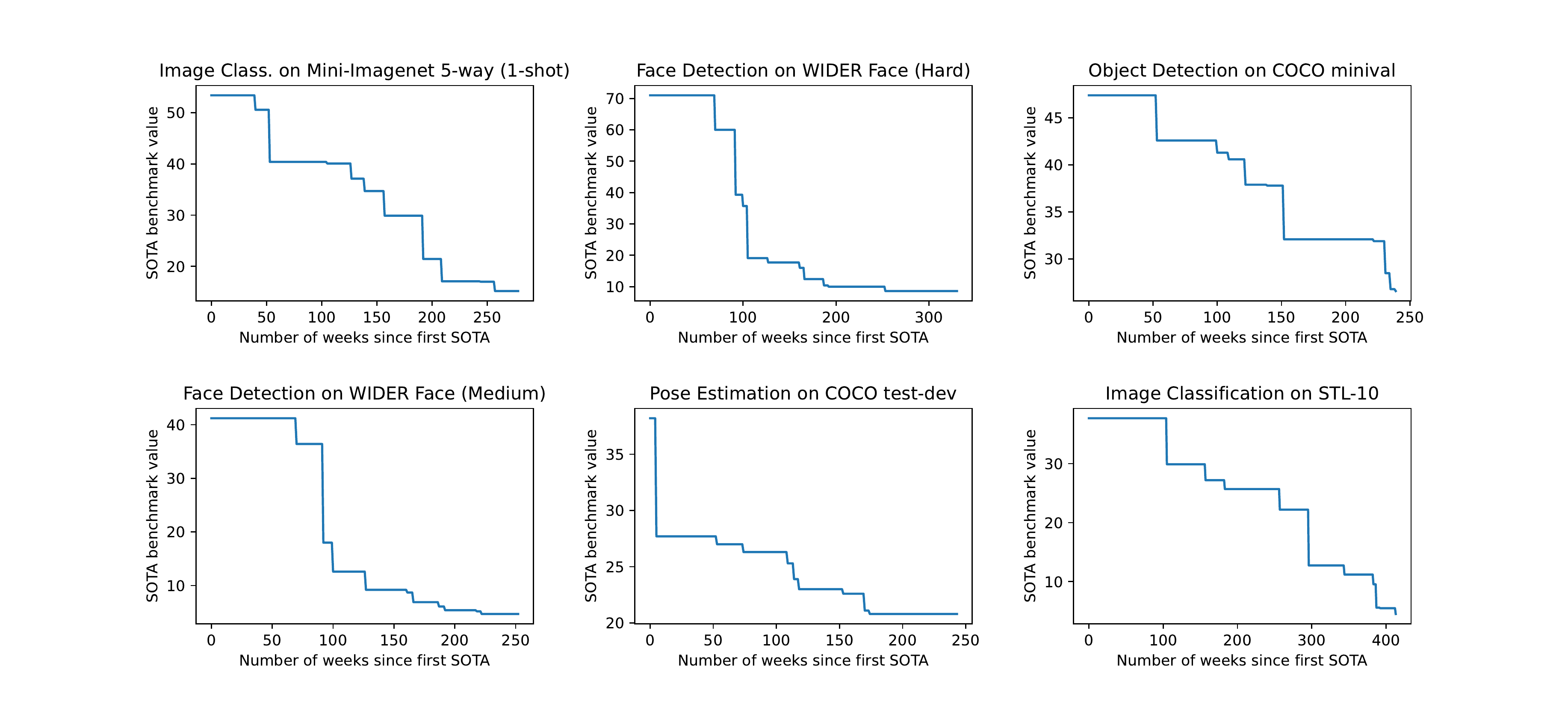}
\caption{\centering \small Normalized state-of-the-art performance on six different ML benchmarks over time. Our dataset spans a total of 435 benchmarks and contains 1552 SOTA improvements.}
\label{fig:ml_data}
\end{figure}

The dataset we have access to contains 1552 different state-of-the-art (SOTA) improvements across 435 different benchmarks (see Figure \ref{fig:ml_data}). 

We use a modified script from \href{https://github.com/skaltman/ai-progress/tree/master/scripts/sota}{Altman 2020} to scrape all Machine Learning results from \href{https://paperswithcode.com/}{Papers with Code} with a cutoff in November 2021. Of these, we select the benchmarks for which performance is reported in accuracy, precision or error-related metrics that are bounded between 0\% and 100\%. \footnote{A full list of the 318 included metrics can be found at \href{https://epochai.org/data/speedrunning}{epochai.org/data/speedrunning}.}. Each metric is associated to multiple benchmarks; we included in our final dataset benchmarks with eight or more records.

Our machine learning data is more limited than our speedrunning data. The benchmark with the most improvements consists of 13 data points (equivalently, 14 distinct SOTA results). Nevertheless, a random effects model may still be applied.\footnote{The data we use for the machine learning benchmark analysis can be downloaded at \href{https://epochai.org/data/speedrunning}{epochai.org/data/speedrunning}.
}


\subsection{Model}

All benchmarks are specified as error rates in the interval \( (0, 1) \) such that smaller values indicate better model performance. As this is a different range than in the speedrunning case, we need a different transformation to put the values in an appropriate space. In this case, we chose the inverse sigmoid transformation and didn't experiment with other choices to avoid overfitting the dataset. The model we fit to the data is then

\[ \log(\sigma^{-1}(R_{c, t}) - \sigma^{-1}(R_{c, t+1})) = \alpha_c + \beta_c \times \log(t) + \varepsilon_{c, t} \]

where \( \sigma^{-1}: p \to \log(p/(1-p)) \) is the inverse sigmoid function.

An important question, which is more difficult to answer here due to the lack of long time series, is whether there's evidence of power-law type decay relative to, e.g., exponential decay in improvements. It's possible that an alternative specification such as

\[ \log(\sigma^{-1}(R_{c, t}) - \sigma^{-1}(R_{c, t+1})) = \alpha_c + \beta_c \times t + \varepsilon_{c, t} \]

would achieve a better fit, and it's worth testing whether this is the case. We'll call this alternative specification ``the random effects model with exponential decay" going forward.

\subsection{Results}

If we include all categories that have at least four SOTA results (equivalently, three SOTA improvements) in our analysis and leave out the final improvement when fitting the random effects model, we get the fit for the random effects model with a power-law decay found in Table \ref{tab:ml-power-law-results}.

\begin{table}[h]
\caption{Mixed Linear Model Regression Result: Machine Learning}
\begin{center}
\begin{tabular}{llll}
\hline
Model:            & MixedLM & Dependent Variable: & \( \log(\sigma^{-1}(R_{c, t}) - \sigma^{-1}(R_{c, t+1})) \) \\
No. Observations: & 967     & Method:             & ML            \\
No. Groups:       & 254     & Scale:              & 1.1615        \\
Min. group size:  & 2       & Log-Likelihood:     & -1506.5608    \\
Max. group size:  & 12      & Converged:          & Yes           \\
Mean group size:  & 3.8     &                     &               \\
\hline
\end{tabular}
\end{center}

\begin{center}
\begin{tabular}{lrrrrrr}
\hline
                          &  Coef. & Std.Err. &       z & P$> |$z$|$ & [0.025 & 0.975]  \\
\hline
\( \mathbb E[\alpha_c] \)                   & -1.545 &    0.065 & -23.796 &       0.000 & -1.672 & -1.418  \\
\( \mathbb E[\beta_c] \)                    & -0.573 &    0.058 &  -9.963 &       0.000 & -0.686 & -0.460  \\
\( \operatorname{var}(\alpha_c) \)          &  0.203 &    0.049 &         &             &        &         \\
\( \operatorname{cov}(\alpha_c, \beta_c) \) &  0.000 &    0.000 &         &             &        &         \\
\( \operatorname{var}(\beta_c) \)           &  0.000 &    0.000 &         &             &        &         \\
\hline
\end{tabular}
\end{center}
\label{tab:ml-power-law-results}
\end{table}

We tried this fit with various masking settings, i.e., various choices of which elements of the covariance matrix of \( \alpha_c, \beta_c \) are forced to vanish. We found that making the variance of \( \alpha_c \) vanish had an adverse effect on the model's goodness of fit and out-of-sample performance, but forcing \( \operatorname{var}(\beta_c) \) or \( \operatorname{cov}(\alpha_c, \beta_c) \) to vanish didn't affect the model's performance and even improved the numerical stability of the optimizer, as the maximum likelihood estimate for the parameters seems to lie near the boundary of the parameter space. As such, the above results are presented with these values forced to zero (or a value such as \( 10^{-6} \) that's practically indistinguishable from zero, to help out the BFGS optimizer).

Table \ref{tab:ml-benchmark-results} compares the error rates of zero-improvment baseline, the power-law decay model and the exponential decay model when we use them to predict the final improvement that we had left out when fitting the model in all categories.

\begin{table}[h]
\centering
\begin{tabular}{@{}lcc@{}}
\toprule
\textbf{Model}                                          & \textbf{Mean \( L^2 \) error} & \textbf{Mean \( L^1 \) error} \\ \midrule
Baseline: always predict zero improvement               & \( 24.04 \% \)                & \( 16.9 \% \)                \\ 
Random effects model, exponential decay                 & \( 17.3 \% \)                & \( 10.8 \% \)                \\
Random effects model, power-law decay                   & \( \bf{16.93 \%} \)                & \( \bf{10.61 \%} \)                \\ \bottomrule
\end{tabular}
\caption{Machine learning benchmark forecasting results.}
\label{tab:ml-benchmark-results}
\end{table}

 Both random effect models have a substantial edge over the naive baseline. If we bootstrap the difference between the mean \( L^2 \) error of the baseline and the power-law decay specifications from the 254 out-of-sample forecasts they make, we get that the power-law decay model beats the baseline at a \( p < 10^{-5} \) significance threshold. The comparison of the exponential decay to the power-law decay model is too noisy to say which model is better. Still, the fact that the difficulties of the examples are correlated across models means that the small differences in mean \( L^2 \) and \( L^1 \) errors in Table \ref{tab:ml-benchmark-results} become statistically significant at quite strict thresholds.

Overall, while we can confidently say that ML benchmarks show diminishing returns, our dataset is inadequate to answer the question of whether returns diminish according to a power-law structure or some other structure. 

It's possible that the power-law decay model is incorrect: the power-law structure in speedrunning benchmarks could be caused by a ``resource overhang" in which the amount of effort going into speedrunning a game is initially far above its difficulty level, suitably measured. It's not clear if ML benchmarks could be expected to differ from speedrunning benchmarks on this point, but if they did, we might not see the same diminishing returns pattern in ML as we do in speedrunning.

A model in which power-law decay dominates until the resource overhang disappears, and then improvements shift to exponential, is theoretically well-justified. A standard law of motion such as \( dS/S \sim -S^{-\phi} I^{\lambda} \) \parencite{Jones1995RD} can produce this result, where \( S \) is the current world record and \( I \) represents resources being expended to improve the record at some instant. However, fitting this law of motion directly to our speedrunning data gives poor results, leading us to doubt that this is the correct explanation for what we observe.

\section{Interpretation of the results}
\label{sec:interpretation}

This section discusses how to interpret the previous sections' empirical findings and estimated model coefficients. There's a separate subsection for each high-level point that we thought was important, and they can be read independently.

In the first subsection, we remark on the structure of diminishing returns in the records and find suggestive evidence that ML benchmarks are far from saturation.

In the second subsection, we look at the variance of the model predictions and show the power-law random-effects model for ML records implies that sudden large improvements are rare, but cannot be ruled out.

\subsection{Diminishing returns structure}

The basic qualitative finding is a power-law trend in the diminishing returns structure of speedrun world record improvements. This power-law trend is apparent in visual inspection of the data, and leveraging it allowed us to produce substantially better forecasts than baseline methods. Furthermore, we've shown it's possible to exploit data parallelism using hierarchical methods such as random-effects models to make forecasts about time series with very few recorded improvements. We also have some limited evidence that power-law decay better describes the diminishing returns structure of machine learning SOTA improvements compared to exponential decay models.

Fitting the random-effects model to the first 50 records in each speedrun category gives the fit found in Table \ref{tab:whole-dataset-results}.

\begin{table}[h]
\caption{Mixed Linear Model Regression Results: Complete Speedrun Dataset}
\label{tab:whole-dataset-results}
\begin{center}
\begin{tabular}{llll}
\hline
Model:            & MixedLM & Dependent Variable: & \( \log \log \left( \frac{R_{c, t}}{R_{c, t+1}} \right) \) \\
No. Observations: & 1225    & Method:             & ML            \\
No. Groups:       & 25      & Scale:              & 1.1402        \\
Min. group size:  & 49      & Log-Likelihood:     & -1854.6215    \\
Max. group size:  & 49      & Converged:          & Yes           \\
Mean group size:  & 49.0    &                     &               \\
\hline
\end{tabular}
\end{center}

\begin{center}
\begin{tabular}{lrrrrrr}
\hline
                          &  Coef. & Std.Err. &       z & P$> |$z$|$ & [0.025 & 0.975]  \\
\hline
\( \mathbb E[\alpha_c] \)                     & -2.484 &    0.147 & -16.874 &       0.000 & -2.773 & -2.196  \\
\( \mathbb E[\beta_c] \)                      & -0.934 &    0.039 & -24.057 &       0.000 & -1.011 & -0.858  \\
\( \operatorname{var}(\alpha_c) \)            &  0.256 &    0.090 &         &             &        &         \\
\( \operatorname{cov}(\alpha_c, \beta_c) \)   &  0.000 &    0.000 &         &             &        &         \\
\( \operatorname{var}(\beta_c) \)             &  0.008 &    0.006 &         &             &        &         \\
\hline
\end{tabular}
\end{center}
\end{table}

The most important fact about the results in Table \ref{tab:whole-dataset-results} is that the coefficients \( \beta_c \) appear to follow a normal distribution with mean \( -0.934 \) and standard deviation around \( 0.09 \). The proximity of the coefficient \( \beta_c \) to \( -1 \) is qualitatively significant: if \( \beta_c < -1 \) then the records \( R_{c, t} \) converge to some strictly positive lower bound from above, while if \( \beta_c \geq -1 \) the records \( R_{c, t} \) converge to zero as \( t \to \infty \). More details can be found in \hyperref[sec:asymptotic-behavior]{Appendix A}.

As the empirical estimates of \( \beta \) for speedruns cluster around \( -1 \), our data doesn't allow us to say with confidence whether the typical speedrun shows sufficiently strong diminishing returns that we can expect the record times to be bottlenecked at a strictly positive greatest lower bound (or infimum). However, we get values of \( \beta \approx -0.5 \) for machine learning benchmarks, which suggests that the diminishing returns trend for these benchmarks is much weaker and shows no sign of being bottlenecked by a lower bound.

We interpret the above finding to mean that ML benchmarks that have been quantified as error rates bounded between \( 0 \) and \( 1 \) are generally far from their optimal possible values, which is why the effect of the strongly diminishing returns that would appear if we were getting close to them doesn't show up in the power-law regressions. These values are also sometimes referred to as \textit{irreducible losses} in the machine learning literature, so empirical performance seems far from the irreducible loss on the benchmarks we considered in this study.

From prior considerations, we suspect that most machine learning benchmarks quantified as error rates should have a strictly positive irreducible loss. Therefore, our findings suggest that we might expect a regime shift as we get closer to these bounds, as the present rates of improvement with \( \beta \approx -0.5 \) don't seem sustainable in such a situation. The important finding is that there doesn't seem to be a sign of such a slowdown in our data.

\subsection{Variance}

While the power-law fit tells us about the \textit{average} decay behavior of world record improvements, we are also interested in the variance: how far and how frequently can we deviate from this trend in the data? For instance, if a power-law fit predicts that our median over the next world record improvement should be \( 1\% \), what should our \( 90 \% \) confidence interval be?

The best way to understand the variance properties of the model is to look at it in log inverse sigmoid space in the ML benchmark case and double log space in the speedrunning world records case, because these spaces are where the models behave homoskedastically. In this case, the standard deviation of the model residuals is around \( 1 \) in both cases, meaning that a \( 95 \% \) confidence interval for improvements at a given time spans a range that's of rough size \( e^4 \approx 54 \) or close to two orders of magnitude wide in relative improvement space. This means the model does not rule out sudden and large improvements in a benchmark.

As a rough example, suppose that the first improvement a benchmark showed was \( 1 \) unit in log odds space. Our model predicts that for improvements of this magnitude to become less than \( 2 \% \) likely, we need the total number of SOTA improvements \( T \) to reach approximately \( e^4 \approx 55 \), at which point the \textit{median} improvement is actually of size \( 0.135 \) units or so. Improvements that are an order of magnitude above the recent past are therefore not ruled out under our model, though they should occur relatively infrequently, e.g., at a rate of once per \( 50 \) improvements or so.

This prediction holds up when we look at the actual distribution of the errors made by the model: roughly \( 2 \% \) of them are off by a factor \( e^2 \approx 7.4 \), which is what we would expect based on the above calculation. The normal distribution assumption seems to be giving good guidance in this case, which is some evidence that our model specification is working as it should.

\section{Conclusion}

In this article, we have studied a remarkable power-law pattern in video game speedrunning record improvements. Exploiting this pattern, we have shown how to build a random effects model of record size improvements. We've shown this model is significantly better than a baseline prediction of no improvement - and arguably better than an exponential moving average fit ex-post to each trend (despite having access to fewer data). We have also shown how to combine this result with a simple model for predicting the timing of records to solve the outstanding problem of predicting speedrunning records eight weeks in advance with better than baseline accuracy.

We then translate those findings into the context of Machine Learning benchmarks. Similarly to the speedrunning case, we show how the power-law decay random effects model can significantly improve on a baseline forecast of no improvement. However, due to the more limited data we cannot discard that another model would be more appropriate, like a random effects model with exponential decay.

The models we built weakly suggest some interesting implications. While in the speedrunning case, the benchmarks are arguably close to saturation, this is not the case for Machine Learning benchmarks. We seem to be far from irreducible losses that would slow down their progress. Also, the model suggests that sudden large improvements in Machine Learning benchmarks are rare but cannot be ruled out. In particular, the model implies that improvements over one order of magnitude in size happen once every fifty times at most.

This foray into the dynamics of record improvements has helped us gather evidence about two key questions of AI forecasting. More work remains to be done to study, among other things, the frequency and size of discontinuities we can expect of the field and the timing of new SOTA improvements.

\setlength\bibitemsep{0.3\baselineskip}
\printbibliography
\newpage
\section{Appendices}

\subsection{Appendix A: Asymptotic behavior of the model}
\label{sec:asymptotic-behavior}

Consider the deterministic model

\[ \log \log \left( \frac{R_{c, t}}{R_{c, t+1}} \right) = \alpha_c + \beta_c \times \log(t) \]

We can approximate this discrete model by a continuous model without affecting the asymptotic properties of the model - this amounts to applying the integral test for the convergence of series in this particular situation. Doing this gives the continuous model

\[ \log \left( -\frac{d \log R_{c, t}}{dt} \right)  = \alpha_c + \beta_c \times \log(t) \]

or

\[ \frac{d \log R_{c, t}}{dt} = -A t^{\beta_c} \]

where \( A = e^{\alpha_c} \) by definition. Integrating from \( t = 1 \) to \( t = T \) gives

\[ \log R_{c, T} = \log R_{c, 1} - \frac{A T^{\beta_c + 1} - A}{\beta_c + 1} \]

where the special case of \( \beta_c = -1 \) is handled using the limit

\[ \lim_{\beta_c \to -1} \frac{A T^{\beta_c + 1} - A}{\beta_c + 1} = A \log T \]

It's apparent that this equation implies

\begin{equation}
    \lim_{T \to \infty} R_{c, T} =
    \begin{cases}
      R_{c, 1} \times \exp \left( \frac{A}{\beta_c + 1} \right) & \text{if $ \beta_c < -1 $} \\
      0        & \text{otherwise}
    \end{cases}
\end{equation}

and so the asymptotic behavior of the model is controlled by whether the parameter \( \beta_c \) is less than \( -1 \) or not. The case \( \beta = -1 \) reduces to a power-law decay in \( R_{c, T} \).

\subsection{Appendix B: Relationship to iterative sampling models}
\label{sec:iterative-sampling}

A tempting way to model time series that show discontinuous progress is to assume that they are the result of \textit{iterative sampling} from some distribution. In other words, we assume that there's some underlying distribution with cumulative distribution function \( F \) that attempts at breaking the current world record are drawn from, and whenever a new sampled attempt is below the cumulative minimum error rate or best time achieved so far, we have a discontinuity in the time series, and the world record moves downward. Formally, we would have

\begin{align}
    X_1, X_2, \ldots, X_k, \ldots &\sim F \\
    Y_i &= \min \{ X_1, X_2, \ldots, X_i \}
\end{align}

and \( Y_i \) would be the world record after \( i \) attempts have been made at the task.

This model has been used previously in the literature. For instance, \cite{kortum1997research} and \cite{jones2021recipes} apply roughly this model of progress to modeling productivity growth in endogenous growth models, while \cite{Sevilla2022} uses a similar model to forecast world records in athletic events.

This model, however, runs into an immediate conflict with data if we assume that the sampling rate is constant. This is because as long as the cumulative distribution function \( F \) is invertible, constant sampling rates should produce exponentially growing time gaps between consecutive world record improvements. The formal proof of this is rather technical and may be found in \hyperref[sec:expon]{Appendix C}, but the intuition is simple: if we have a record from the top \( 10 \% \) of the distribution, it'll take us on average \( 10 \) samples from the distribution to break this record. Once the record has been broken, we'll end up with a new record around the top \( 5 \% \) range on average, so for the next record improvement we'll need \( 20 \) more samples instead. Imagining how this process works should be sufficient to provide intuition for why the time gaps should be exponential.

We don't see this in the data: the time gaps between consecutive record improvements appear to be either constant or at best linearly increasing at a slow pace, while the exponential time gaps result predicts, not only that we should see exponential growth, but that the growth should be roughly by a factor of \( e \) between each consecutive record improvement. This result is decisively rejected by both the speedrunning and the machine learning benchmark data without any statistical testing: it's enough to just look at the time series to see that this is true.

\cite{jones2021recipes} overcomes this problem by assuming that what we're searching for are \textit{recipes}, and assumes that when we have \( N \) recipes we have an effective number of \( 2^N \) subsets of these recipes that we can combine to form an ``attempt." This is equivalent to changing the sampling rate in the above setup to grow exponentially, and in this case we asymptotically get a constant time between consecutive world record improvements on average. However, we think this is not the right way to think about how record improvements actually work: It's possible to fit \textit{any} monotone discontinuous time series by freely varying both the tail behavior of the distribution \( F \) and the sampling rate intensity that's being used to sample from \( F \), so the idea of using iterative sampling loses much of its explanatory power when we drop the constant sampling rate assumption.

Still, we might want to investigate which distribution actually does reproduce the power-law type decay models that we've used above. This distribution needs to have the property that its quantile function \( Q(p) \) satisfies

\[ \frac{Q(e^{-t})}{Q(e^{-t-1})} \approx e^{e^{\alpha + \beta \times \log t}} = e^{e^{\alpha} t^{\beta}} \]

Iterating backwards gives

\[ Q(e^{-t}) \approx Q(e^{-1}) \times \prod_{k=1}^{t-1} e^{-e^{\alpha} k^{\beta}} \approx Q(e^{-1}) \times \exp \left( -\frac{e^{\alpha} (t^{\beta+1} - 1)}{\beta+1} \right) \]

or, upon reparametrization,

\[ Q(p) \approx  Q(e^{-1}) \times \exp \left( -\frac{e^{\alpha} ((- \log p)^{\beta+1} - 1)}{\beta+1} \right) \]

As \( \beta \to -1 \), this reduces to

\[ Q(p) \approx Q(e^{-1}) e^{-e^{\alpha} \log(\log(1/p))} = Q(e^{-1}) (-\log p)^{-e^{\alpha}} \]

This distribution is the \textit{exponential} of the Gumbel distribution. In other words, our model implies that \( \beta = -1 \) matches the results we would expect from an iterative sampling model if the logarithms of world records were being iteratively sampled from a Gumbel distribution. The power-law type decay model presented here, therefore, has Gumbel distribution type decay as a special case. It's possible that the findings in \cite{Sevilla2022} need to be revised in the face of this finding, as they assume that the raw records \( R_{c, t} \) and \textit{not} their logarithms are being sampled from a Gumbel distribution, potentially affecting their results significantly.

\subsection{Appendix C: Exponential time gaps}
\label{sec:expon}

One of the basic results in optimization by random search is that the time gaps between successive optimal values, or records, obtained so far grow exponentially under a suitable regularity condition on the function being optimized.

Formally, if we have a measurable function \( f: \Omega \to \mathbb R \) from a probability space \( \Omega \) to the real numbers, we can interpret it as a random variable and attempt to maximize it by sampling successively from the probability distribution on \( \Omega \) and evaluating \( f \) at those points. In the context of this article, we'll work with trying to minimize \( f \) using random sampling, though a maximizer works just the same. As optimization by random search only sees the order structure between the values of \( f \), the optimization process will not change if we compose \( f \) by a strictly increasing function.

Assume that the law of \( f \), or the cumulative distribution function \( F(x) = \mathbb P(f \leq x) \) is invertible, so that \( F \) actually defines a bijection \( \mathbb R \to (0, 1) \). Composing \( f \) with \( F \) gives a new function \( F \circ f: \Omega \to [0, 1] \) which has the property that its law is uniformly distributed: the push-forward of the measure on \( \Omega \) to \( [0, 1] \) is simply the uniform distribution. A direct computation is sufficient to show this:

\[ \mathbb P(F \circ f \leq x) = \mathbb P(f \leq F^{-1}(x)) = F(F^{-1}(x)) = x \]
    
which is the cumulative distribution function of the uniform distribution. This transformation corresponds to looking at the percentiles of the values of \( f \) rather than the absolute levels. So without loss of generality we may assume \( f \) itself has codomain \( [0, 1] \) and the distribution it induces on \( [0, 1] \) is uniform.

We can now prove the exponential time gaps hypothesis:

\begin{theorem}[Exponential record times]
\label{exponential_record_times}
If \( N_1, N_2, \ldots \) are the sequence of \textit{record times} during the minimization process of \( f: \Omega \to \mathbb R \) by random rejection sampling, and \( f \) defines an invertible cumulative distribution function on \( \mathbb R \), then \( \lim_{i \to \infty} (N_i)^{1/i} = e \) almost surely.

A \textit{record time} is defined as follows: if an optimizer evaluates \( f \) at the sequence of points \( p_1, p_2, p_3, \ldots \), then a \textit{record time} is an index \( i \) such that \( f(p_i) < f(p_j) \) for all \( j < i \).
\end{theorem}

This theorem states that \textit{on average} the time gaps between new records increase by a factor of \( e \) at each step, and therefore so do the gaps between successive records. This justifies the maxim of ``exponential time gaps."

\subsubsection{Sketch of proof}

The probability mass function of \( N_{i+1} - N_i \) conditional on \( N_i \) is given by

\[ k \to N_i \int_0^1 (1-x) x^{N_i+k-2} \, dx = \frac{N_i}{(N_i + k)(N_i + k - 1)} = \frac{N_i}{N_i + k - 1} - \frac{N_i}{N_i + k} \]

We can use this to compute \( \mathbb E[\log(N_{i+1}/N_i) \vert N_i] \) as

\[ \mathbb E[\log(N_{i+1}/N_i) \vert N_i] = \frac{1}{N_i} \sum_{k=1}^{\infty} \frac{\log(1 + k/N_i)}{(1 + k/N_i)(1 + (k-1)/N_i)} \]

For \( N_i \) large, this sum is well approximated by the integral

\[ \int_0^{\infty} \frac{\log(1+x)}{(1+x)^2} \, dx = 1 \]

and the error in this approximation is \( O(1/N_i) \). Moreover, the second moment is also obviously \( O(1) \), and therefore so is the variance. It follows that

\[ \mathbb E[\log(N_i)] = \mathbb E[\log(N_1)] + \sum_{h=1}^{i-1} \mathbb E[\log(N_{h+1}/N_h)] = \mathbb E[\log(N_1)] + (i-1) + O(1) \]

\[ \sigma^2(\log(N_i)) = O(i) \]

after some routine computations using the fact that the variables \( \log(N_{i+1}/N_i) \) are almost orthogonal (the covariances are all of order \( 1/N_i \)) for distinct values of \( i \) when the \( N_i \) are large, as in this situation the distribution of \( \log(N_{i+1}/N_i) \) is close to being independent of the value of \( N_i \) as seen in the computation of the expected value above. Therefore Chebyshev's inequality suffices to prove the convergence result in the theorem in the sense of convergence in probability after taking exponentials.

More explicitly, for \( i > j \):

\begin{align}
\mathbb E[\log(N_{i+1}/N_i) \log(N_{j+1}/N_j))] &= \mathbb E[\mathbb E[\log(N_{i+1}/N_i) \vert N_1, \ldots, N_i] \log(N_{j+1}/N_j))] \\
&= \mathbb E[(1 + O(1/N_i)) \log(N_{j+1}/N_j)] \\
&= 1 + O(\mathbb E[1/N_j])
= 1 + e^{-O(j)}
\end{align}

so that

\[ \operatorname{cov}(\log(N_{i+1}/N_i), \log(N_{j+1}/N_j)) = e^{-O(j)} \]

This gives, with the convention that \( N_0 = 1 \),

\begin{align}
\sigma^2(\log(N_i)) &= \sigma^2 \left(\sum_{h=0}^{i-1} \log(N_{h+1}/N_h)) \right) \\ 
&= \sum_{0 \leq h_1, h_2 \leq i-1} \operatorname{cov}(\log(N_{h_1+1}/N_{h_1})), \log(N_{h_2+1}/N_{h_2}))) \\
&= \sum_{0 \leq h_1, h_2 \leq i-1} e^{-O(\min\{h_1, h_2 \})} \\
&\leq \sum_{0 \leq h \leq i-1} i e^{-O(h)}
= O(i)
\end{align}

which finishes the proof of the convergence in probability.

To strengthen this to almost sure convergence, we can employ the fourth moment tail bound in a similar fashion to the proof of the strong law of large numbers for random variables of finite fourth moment. The proof is relatively straightforward, with the only technical detail being about how the \( \log(N_{i+1}/N_i) \) are not quite independent for distinct \( i \). However, as the error made by approximating the expected values of well-behaved functions of \( \log(N_{i+1}/N_i) \) by corresponding integrals is of order \( \sim 1/N_i \), which is exponential in \( \log(N_i) \), the failure of the increments to be exactly orthogonal poses no obstacle to the proof.

\end{document}